\documentclass[10pt,twocolumn,letterpaper]{article}

\usepackage{cvpr}              

\usepackage{graphicx}
\usepackage{amsmath}
\usepackage{amssymb}
\usepackage{booktabs}
\usepackage{multirow}
\usepackage{bbding}
\usepackage{balance}

\usepackage{fontawesome5}
\usepackage{newfloat}
\usepackage{listings}
\usepackage{makecell}
\usepackage{colortbl}
\usepackage{xcolor}
\usepackage{algorithm}
\usepackage{algorithmic}
\usepackage{pifont}

\newtheorem{proposition}{Proposition}

\definecolor{cvprblue}{rgb}{0.21,0.49,0.74}
\usepackage[pagebackref,breaklinks,colorlinks,allcolors=cvprblue]{hyperref}

\usepackage[capitalize]{cleveref}
\crefname{section}{Sec.}{Secs.}
\Crefname{section}{Section}{Sections}
\Crefname{table}{Table}{Tables}
\crefname{table}{Tab.}{Tabs.}

\newcommand\ies{\textit{i.e.}}

\def\paperID{3254} 
\def\confName{CVPR}
\def\confYear{2026}

\begin{document}


\title{Multimodal Learning on Low-Quality Data with Conformal Predictive Self-Calibration}

\author{Xun Jiang${^{1,2}}$, Yufan Gu${^{2}}$, Disen Hu${^{2}}$, Yuqing Hou${^{3}}$, Yazhou Yao${^{4}}$, Fumin Shen${^{2}}$, \\
Heng Tao Shen${^{1}}$, Xing Xu${^{1}}$\thanks{Corresponding author.} \\
{$^1$School of Computer Science and Technology, Tongji University}\\
{$^2$School of Computer Science and Engineering, }\\{ University of Electronic Science and Technology of China}, {$^3$Independent Researcher} \\
{$^4$School of Computer Science and Engineering, Nanjing University of Science and Technology}\\
}

\maketitle

\begin{abstract}
  Multimodal learning often grapples with the challenge of low-quality data, which predominantly manifests as two facets: modality imbalance and noisy corruption. 
  While these issues are often studied in isolation, we argue that they share a common root in the predictive uncertainty towards the reliability of individual modalities and instances during learning. 
  In this paper, we propose a unified framework, termed \textbf{C}onformal \textbf{P}redictive \textbf{S}elf-\textbf{C}alibration (\textbf{CPSC}), which leverages conformal prediction to equip the model with the ability to perform self-guided calibration on-the-fly. 
  The core of our proposed CPSC lies in a novel self-calibrating training loop that seamlessly integrates two key modules: 
  (1) Representation Self-Calibration, which decomposes unimodal features into components, selectively fuses the most robust ones identified by a conformal predictor to enhance feature resilience. 
  (2) Gradient Self-Calibration, which recalibrates the gradient flow during backpropagation based on instance-wise reliability scores, steering the optimization towards more trustworthy directions. 
  Furthermore, we also devise a self-update strategy for the conformal predictor to ensure the entire system co-evolves consistently throughout the training process. 
  Extensive experiments on six benchmark datasets under both imbalanced and noisy settings demonstrate that our CPSC framework consistently outperforms existing state-of-the-art methods. 
  Our code is available at \url{https://github.com/XunCHN/CPSC}.
\end{abstract}

\section{Introduction}
\label{sec:intro}

\begin{figure}[thb]
    \centering
    \includegraphics[width=0.98\linewidth]{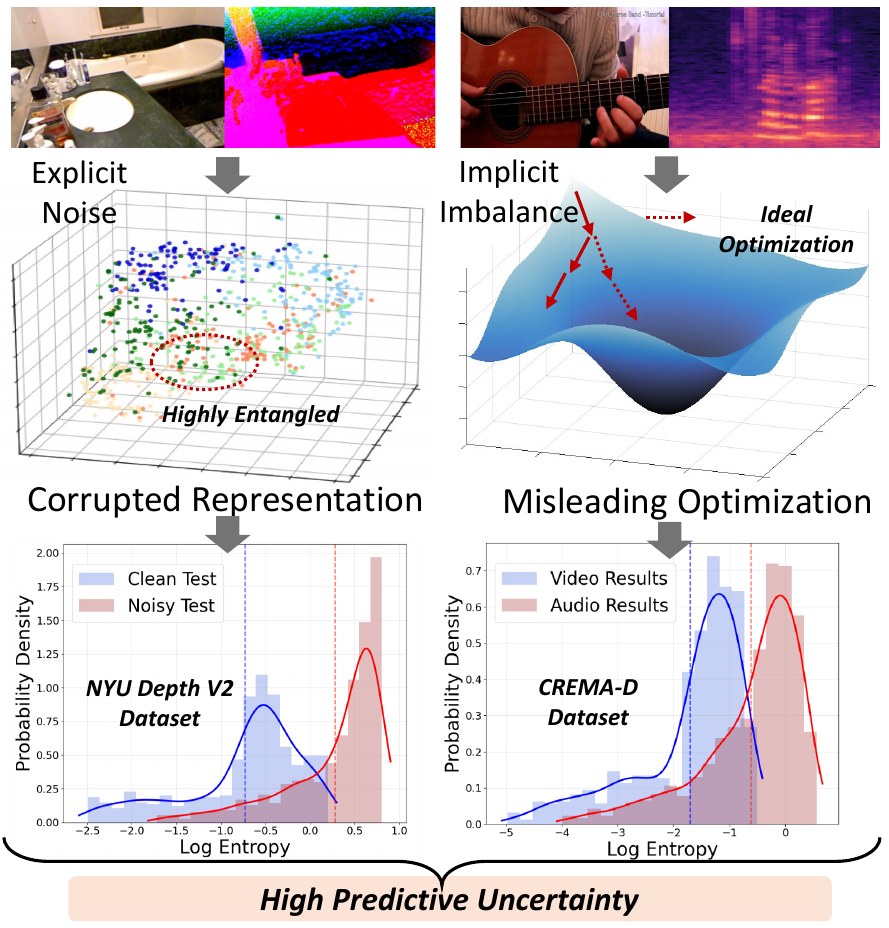}
    \caption{Implicit imbalance and explicit noise corruption would increase the model predictive uncertainty. Here we illustrate the statistical log entropy of predicted distributions on NYU Depth V2 \cite{silberman2012indoor} and CREMA-D \cite{cao2014crema} datasets with LFM method \cite{LFM_nips2024}. }
    \label{fig:intro}
    \vspace{-4mm}
\end{figure}

The proliferation of multimodal sensors has catalyzed remarkable progress in multimodal learning \cite{ml_1, ml_2,xun_JOSFD}. 
However, practical systems frequently grapple with low-quality data \cite{rml_1, EAU_CVPR2024, iml_1}, which primarily manifests in two detrimental forms: 
(1) Implicit modality imbalance \cite{iml_1, iml_2} with skewed data distributions, causing the model to be biased towards dominant modalities. (2) Explicit noisy corruption \cite{QMF_ICML23,EAU_CVPR2024}, where one or more modalities are contaminated by dynamic noise, misleading the learning process. 

Traditionally, these two challenges have been addressed in isolation. 
A suite of techniques, such as weighted losses \cite{iml_3,GGDM_MM2025}, gradient modulation \cite{iml_1, iml_2}, and data re-sampling \cite{iml_4, iml_5}, has been developed to handle modality imbalance. 
Meanwhile, another line of research focuses on learning from noisy modalities via robust fusion \cite{EAU_CVPR2024, QMF_ICML23, rml_4}. 
While effective within their specific domains, this divided approach overlooks a fundamental commonality: 
As illustrated in Fig. \ref{fig:intro}, both imbalance and noise exacerbate models' predictive uncertainty, which is defined as the model's inherent doubt about the correct prediction for a given sample. 
Imbalance leads to under-represented modalities being ignored, while noise injects misinformation.
Both of these result in overconfident but unreliable predictions. 
Consequently, the lack of a unified framework that explicitly quantifies and mitigates model uncertainty limiting the generalization and robustness of existing methods.

\begin{figure*}[t]
    \centering
    \includegraphics[width=0.97\linewidth]{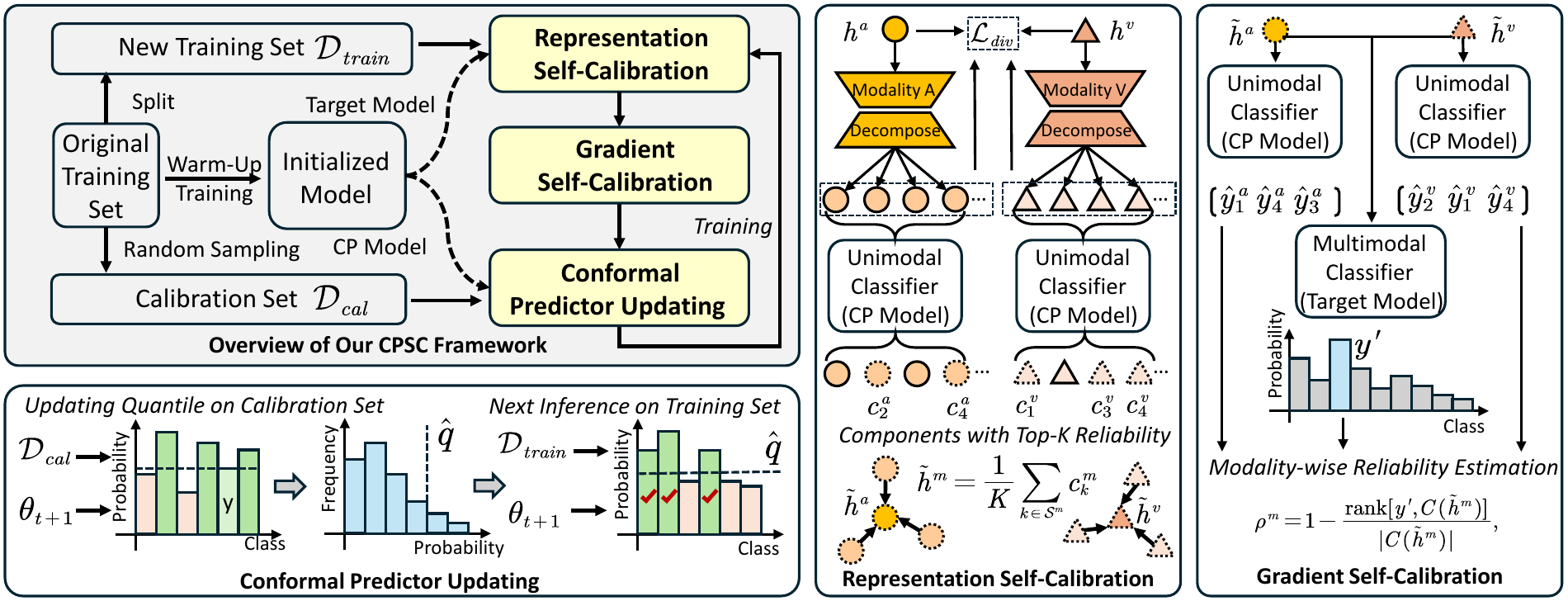}
    \caption{The overall architecture of our CPSC framework, illustrating the self-calibration training loop with Representation Self-Calibration (RSC), Gradient Self-Calibration (GSC), and Conformal Predictor Updating.}
    \label{fig:framework}
    \vspace{-4mm}
\end{figure*}

Following this observation, we argue that multimodal learning on low-quality data can be unified by addressing its root cause, \ies, insufficient awareness and calibration of predictive uncertainty. 
However, as multimodal data can be corrupted by both explicit and implicit factors, existing methods are specialized for particular settings. 
Hence, \textit{how to devise a model-agnostic, unified predictive uncertainty quantification approach} becomes an essential problem.
To address this problem, we draw inspiration from Conformal Prediction (CP) \cite{cp_1}, a statistical framework providing prediction sets with finite-sample and distribution-free guarantees. 
Unlike Bayesian methods \cite{bayesian_1, bayesian_2} that require priors, CP offers a model-agnostic way to quantify predictive uncertainty. For multimodal learning on low-quality data, we propose a novel unified framework termed \textit{Conformal Predictive Self-Calibration} (\textit{CPSC}). 
As illustrated in Fig. \ref{fig:framework}, CPSC leverages a dynamically maintained CP model to instill a self-calibration mechanism into the training loop. 
It enables the model to continuously diagnose its uncertainty and correctify its learning trajectory.

The core of our CPSC framework is a novel training paradigm that integrates three key innovations:
(1) \textit{Representation Self-Calibration} that decomposes unimodal features into orthogonal components and employs the CP model to identify and retain the most reliable components for fusion. 
(2) \textit{Gradient Self-Calibration} that uses the CP model to recalibrate the gradient during backpropagation, steering the optimization towards more trustworthy directions and mitigating the influence of ambiguous or corrupted samples.
(3) \textit{Conformal Predictor Updating} strategy where the model parameters and the CP model co-evolve by refreshing the CP model with optimized model.
Extensive experiments on six benchmark datasets under both imbalanced and noisy settings demonstrate that our proposed CPSC consistently outperforms state-of-the-art methods. 
Overall, our contributions in this work are threefold:
\begin{itemize}
  \item We propose the CPSC framework, which addresses the multimodal learning from a unified and model-agnostic perspective, linking modality imbalance and noisy corruption through the lens of model predictive uncertainty. 
  \item We propose the Representation Self-Calibration and Gradient Self-Calibration modules, which effectively integrates Conformal Prediction into the core of multimodal training, introducing self-calibration mechanisms for feature representation and gradient optimization.
  \item We provide extensive empirical and theoretical analysis, showing that CPSC achieves new state-of-the-art performance across diverse low-quality data scenarios.
\end{itemize}

\section{Related Work}
\label{sec:rel}

\noindent
\textbf{Multimodal Learning on Low-Quality Data.}
Practical multimodal learning often contends with imperfect data \cite{xun_ART, wang_TPAMI25, xun_GETV}, which has been primarily studied on the imbalanced and noisy multimodal data. 
Early methods often employed static fusion strategies, such as weighted averaging \cite{iml_8} or gated mechanisms \cite{iml_9}. 
Recent approaches dynamically modulate the influence of each modality. 
For instance, some methods \cite{iml_1, InfoReg_CVPR2025, GGDM_MM2025, Mmpareto_ICML24} use gradient manipulation to balance the learning pace across modalities, preventing one from dominating the training. 
Meanwhile, other researchers aim to improve the multimodal learning robustness towards dynamic noise. 
In this paradigm, a common strategy is to design robust fusion architectures that can selectively focus on reliable information \cite{EAU_CVPR2024, rml_4}. 
Other methods \cite{TMC_TPAMI22, rml_6} explicitly model the reliability of modalities to down-weight untrustworthy signals during fusion. 
However, existing works above are developed as specialized solutions, thus lacking a unified mechanism to address these low-quality multimodal learning jointly.

\noindent
\textbf{Predictive Uncertainty.} 
Predictive uncertainty is a crucial concept for reliable machine learning, which quantifies models' lack of confidence in their outputs for a given input. 
It has been widely used in massive computer vision or multimodal tasks, such as video understanding \cite{xun_NREP, xun_JOSFD}, cross-modal retrieval \cite{shenshen_tcsvt24, shenshen_aaai24}, and autonomous driving \cite{pu_1, pu_2, pu_8}. 
In multimodal learning, predictive uncertainty estimation has also been prevalent to improve fusion robustness. 
Typically, a group of methods \cite{hu2026hyper, TMC_TPAMI22, pu_8, EAU_CVPR2024, pu_7} use uncertainty to weight modalities in late fusion \cite{TMC_TPAMI22, pu_8} or to guide the fusion process in intermediate layers \cite{EAU_CVPR2024, pu_7}. 
However, these approaches relied on specified model designs with specific architectural choices or prior assumptions. 
In contrast, our proposed CPSC approach employs conformal prediction, which address the predictive uncertainty estimation in a model-agnostic and distribution-free manner. 

\noindent
\textbf{Conformal Prediction.}
The CP \cite{cp_1} is a statistical framework that produces prediction sets with guaranteed coverage probabilities, meaning the true label is contained within the set with a user-specified probability. 
Its model-agnostic nature has led to applications in risk control \cite{cp_risk_1}, anomaly detection \cite{cp_ano_1, cp_ano_2}, large language models \cite{cp_llm_1, cp_llm_2}. 
While CP has seen growing diverse applications, its application in multimodal learning theory remains nascent. 
Existing works have primarily applied CP in a post-hoc manner, for instance, to create predictive sets for multimodal outputs after training is completed \cite{cp_2, cp_3, cp_4}. 
A few works \cite{cp_6, cp_7, cp_8} have begun to integrate CP into the training process, such as loss function. 
Compared to these works, our proposed CPSC pioneeringly builds a self-calibrating training framework that dynamically adjusts both feature-level representation and optimization-level gradients.

\section{Our Proposed Method}
\label{sec:method}

\subsection{Preliminary}
\label{subsec:pre}
The Conformal Prediction \cite{cp_1} is a distribution-free framework that provides predictive sets with statistical guarantees. 
Consider a classification task with label space $\mathcal{Y} = \{1, 2, ..., K\}$. 
Given a trained model $f$ and a calibration set $\mathcal{D}_{cal} = \{(x_i, y_i)\}_{i=1}^{n}$ drawn from the same distribution as the test data, CP quantifies the model's uncertainty for a new test sample $x_{test}$.
The core of CP is the nonconformity score $s(x, y)$, which measures how ``strange'' the pair $(x, y)$ is relative to the model's predictions. 
Let $f(x)_y$ represent the predicted probability for the true label $y$, the nonconformity score can be calculated as: 
\begin{equation}
    s(x, y) = 1 - f(x)_y ,
    \label{eq:conformal_score}
\end{equation}
Hence, for a predefined significance level $\alpha \in (0, 1)$, \ies, risk factor, the conformal prediction set is constructed as:
\begin{equation}
    C(x_{test}) = \{y \in \mathcal{Y}: s(x_{test}, y) \leq \hat{q}\}, 
\end{equation}
where $\hat{q}$ is the $\lceil (n+1)(1-\alpha) \rceil / n$-th quantile of the nonconformity scores $\{s(x_i, y_i)\}_{i=1}^{n}$ on the calibration set. 
In this way, the predictions of model on test cases can be guaranteed within marginal coverage:
\begin{equation}
    \mathbb{P}(y_{test} \in C(x_{test})) \geq 1 - \alpha.
\end{equation}

\subsection{Conformal Predictor Construction}

\noindent
\textbf{An Overview of Our CPSC Method.}
As illustrated in Fig. \ref{fig:framework}, our CPSC introduces a self-calibration mechanism into the standard multimodal training pipeline. 
Given a multimodal training dataset $\mathcal{D} = \{(x^m_i, y_i)\}_{i=1}^{N}$ with $M$ modalities, we first split it into two sets randomly with the indentical distributions, \ies, training set $\mathcal{D}_{train}$ and calibration set $\mathcal{D}_{cal}$. 
The framework operates through the following two key phases: \textit{Warm-up Training Phase} for initial training of the multimodal model $f_{\theta}$ on $\mathcal{D}_{train}$ and \textit{Self-Calibration Training Loop}. The overall framework creates a closed-loop system where the model continuously self-calibrates with its evolving understanding of predictive uncertainty. 

Specifically, in the self-calibration training loop, for each training iteration $t$, the training loop follows: (1) Extract unimodal features and apply \textit{Representation Self-Calibration (RSC)} using the current CP model. (2) Fuse calibrated features and compute predictions. (3) Apply \textit{Gradient Self-Calibration} using CP reliability scores before parameter update. (4) Update model parameters $\theta_t \rightarrow \theta_{t+1}$. (5) Execute the Conformal Predictor Updating to refresh the CP model for the next iteration.

\noindent
\textbf{Warm-up Training Phase.}
Before starting the self-calibration training process, we conduct warm-up training process to initialize the CP model. 
Let $f_{\theta} = \{E^1_{\theta_1}, E^2_{\theta_2}, ..., E^M_{\theta_M}, F_{\theta_f}\}$ denote our multimodal model, where $E^m$ are modality-specific encoders and $F$ is the fusion classifier, the overall training objective is:
\begin{equation}
    \theta = \arg\min_{\theta} \mathbb{E}_{(x,y) \sim \mathcal{D}_{train}}[\mathcal{L}_{CE}(f_{\theta}(x), y)],
\end{equation}
where $\mathcal{L}_{CE}$ is the standard cross-entropy loss and $\theta$ is the model parameters. 
We set the warm-up epoch to $t_0$, and the initialized CP model parameters can be denoted as $\theta_{t_0}$
After warm-up, we further initialize the CP model using the calibration set $\mathcal{D}_{cal}$. 
For each $(x_i, y_i) \in \mathcal{D}_{cal}$, we compute the nonconformity score $s_i = s(x_i, y_i)$ using Eq. \ref{eq:conformal_score} and obtain the initial quantile $\hat{q}_{t_0}$.

\subsection{Representation Self-Calibration}
\label{subsec:RSC}

\noindent
\textbf{Feature Decomposition.} 
To enable fine-grained reliability assessment, we first decompose the original feature into multiple components. 
Through the decomposition, we can examine different aspects of the representation separately and identify which components contribute most to reliable predictions. 
Specifically, we first project the original feature $h^m$ into a higher-dimensional space using a modality-specific fully-connected layer $W_{dec}^m \in \mathbb{R}^{l \times d}$ where $l = n \times d$, followed by a ReLU activation:
\begin{equation}
    h_{high}^m = \text{ReLU}({W_{dec}^m} h^m) \in \mathbb{R}^{l \times 1}.
\end{equation}
The high-dimensional feature is then split into $n$ components $\{c^m_k\}_{k=1}^n$, where each component $c^m_k \in \mathbb{R}^d$:
\begin{equation}
    \{c^m_1, c^m_2, ..., c^m_n\} = \text{Split}(h_{high}^m).
\end{equation}
$c^m_k \in \mathbb{R}^{d}$ is considered as a feature component that captures different aspects of the original representation.

To ensure the components are both representative and diverse, we impose constraints using KL divergence:
\begin{equation}
    \begin{aligned}
        \mathcal{L}_{div}^m = & \frac{\lambda_1}{n}\sum_{k=1}^n D_{KL}(P(h^m) \| P(c^m_k)) \\
        ~~~~~ & - \frac{\lambda_2}{n(n-1)}\sum_{i \neq j} D_{KL}(P(c^m_i) \| P(c^m_j)),
    \end{aligned}
    \label{eq:kl_constraint}
\end{equation}
where $P(\cdot)$ denotes the probability distribution obtained by applying softmax to the feature vector. 
$\lambda_1$ and $\lambda_2$ are balance factors. 
By applying the training objective above, we encourage each component to be close to the original feature while maintaining diversity among components.

\noindent
\textbf{Reliability Scoring.} 
Not all feature components contribute equally to reliable predictions, especially in the presence of noisy or imbalanced data. 
Hence, we leverage conformal prediction to quantitatively assess the reliability of each component, providing a principled way to distinguish between robust and noisy features with statistical guarantees.
For each component $c^m_k$, we compute its reliability score using the current CP model. 
Since the CP model shares parameters with the current main model $f_{\theta_t}$ at iteration $t$, we directly feed each component to the corresponding unimodal classifier $F$ to obtain classification probabilities.
Specifically, for component $c^m_k$, we compute:
\begin{equation}
    p^m_k = F_{m}(c^m_k) \in \mathbb{R}^{|\mathcal{Y}|},
\end{equation}
where $p^{m}_k$ represents the predicted probability distribution with modality $m$ over classes. 
We then calculate the nonconformity scores for all classes using Eq. \ref{eq:conformal_score}:
\begin{equation}
    s(c^m_k, y) = 1 - p^m_k[y], \quad \forall y \in \mathcal{Y}.
\end{equation}
The conformal prediction set for component $c^m_k$ follows:
\begin{equation}
    C(c^m_k) = \{y \in \mathcal{Y}: s(c^m_k, y) \leq \hat{q}_t\},
\end{equation}
where $\hat{q}_t$ is the current conformal quantile obtained from the calibration set (Please refer to Sec. \ref{subsec:pre}).
The reliability score $r^m_k$ for component $c^m_k$ is then computed based on the position of the ground truth $y$ in the sorted prediction set:
\begin{equation}
    r^m_k = 1 - \frac{\text{rank}\left[y, C(c^m_k)\right]}{|C(c^m_k)|}. 
    \label{eq:reliability_score}
\end{equation}
$\text{rank}\left[y, C(c^m_k)\right]$ denotes the rank position of the true label $y$ when the labels in $C(c^m_k)$ are sorted by their nonconformity scores in ascending order, and $r^m_k = 0$ if $y \notin C(c^m_k)$.

\noindent
\textbf{Robust Feature Reconstruction.} 
After identifying the most reliable components, we need to reconstruct a robust feature representation that emphasizes these components while suppressing less reliable ones. 
Specifically, we sort all components $\{c^m_k\}_{k=1}^n$ by their reliability scores $\{r^m_k\}_{k=1}^n$ in descending order and select the top-$K$ components with the highest scores. 
Let $\mathcal{S}^m$ denote the set of indices of these selected components,
the calibrated feature $\tilde{h}^m$ is obtained by averaging the selected components:
\begin{equation}
    \tilde{h}^m = \frac{1}{K} \sum_{k \in \mathcal{S}^m} c^m_k.
    \label{eq:robust_reconstruction}
\end{equation}
Here we give a formulated proposition as follows: 
\begin{proposition}
The expected deviation between the calibrated representation $\tilde{h}^m$ and an ideal robust representation $h^m_*$ is bounded by:
\begin{equation}
    \mathbb{E}[\|\tilde{h}^m - h^m_*\|_2] \leq \frac{1}{K} \sum_{k \in \mathcal{S}^m} \mathbb{E}[\|c^m_k - h^m_*\|_2],
\end{equation}
where the selection of components into $\mathcal{S}^m$ ensures that $\mathbb{E}[\|c^m_k - h^m_*\|_2]$ is minimized for $k \in \mathcal{S}^m$.
\end{proposition}
The theorem demonstrate the effectiveness of our calibrated representation, and the proof is available in our supplementary materials. 
Hence, the overall training objective of our CPSC framework comprises of conventional classification loss and diversity loss for RSC across all modalities $\mathcal{L} = \mathcal{L}^\text{mul}_\text{CE} + \sum_{m=1}^M \mathcal{L}_{div}^m$.

\subsection{Gradient Self-Calibration}
After obtaining the final prediction $\hat{y} = F(\{\tilde{h}^m\}_{m=1}^M)$, we compute the multimodal cross-entropy loss $\mathcal{L}_{CE}(\hat{y}, y)$ and unimodal $\mathcal{L}^{m}_{CE}(\hat{y}^{m}, y)$ synchronously. 
Before backpropagation, we apply gradient calibration to guide the optimization process toward more trustworthy directions. 

\noindent
\textbf{Modality-wise Reliability Estimation.} 
To enable modality-wise gradient calibration, we first need to quantify the reliability of each training instance across different modalities. 
By leveraging the conformal prediction framework, we obtain statistically grounded reliability scores that reflect how well each sample aligns with the model's current understanding of multimodal synergy. 
Specifically, for each training sample with multiple modalities in the current batch $\mathcal{B}$, we compute the reliability score of the outputs of each unimodal classifier.
Rather than using the ground truth label $y$, we regard the multimodal predicted label $y^\prime$ as a synergistic recognition of the current model. 
Following that, we conduct a similar process with current CP model inference in \textit{Reliability Scoring} (See Sec. \ref{subsec:RSC}). 
Finally, we obtain the unimodal reliability $\rho^{m}$ that describes the synergistic discrepancy of $m$ modality: 
\begin{equation}
    \rho^m = 1 - \frac{\text{rank}(y^{\prime}, C(\tilde{h}^m))}{|C(\tilde{h}^m)|},
\end{equation}

\noindent
\textbf{Gradient Re-weighting.} 
Once we have quantified the reliability of each modality, we can modulate their contributions to the gradient updates for this sample. 
Specifically, we modulate the loss from each modality based on its reliability using a linear weighting $w(\rho^{m}) = a \cdot \rho^{m} + b$, where $a$ and $b$ are hyperparameters that control the intensity and baseline of the gradient calibration. 
The gradient for parameter $\theta$ with respect to the calibrated modality becomes:
\begin{equation}
    \nabla_{\theta} \mathcal{L}^{m}_{GSC} = \frac{1}{|\mathcal{B}|} \sum_{i=1}^{|\mathcal{B}|} w(\rho^{m}) \cdot \nabla_{\theta} \mathcal{L}^{m}_{CE}(\hat{y}^{m}, y),
\end{equation}
where $\hat{y}$ is the predicted classification results. 
Here we give a formulated proposition to show the GSC module is stable: 
\begin{proposition}
For a convex loss function $\mathcal{L}$, the GSC module with linear weighting reduces the effective variance of the stochastic gradient estimate when $w(\rho)$ is positively correlated with $\|\nabla \mathcal{L}\|_2$. 
\end{proposition}
The proof can be available in our supplementary materials. 
Hence, our re-weighting mechanism reduces the influence of unreliable samples with low $\rho^{m}$ during optimization while amplifying the contribution of highly reliable samples, effectively steering the model toward more confident and accurate predictions.

\begin{algorithm}[thb]
\caption{Self-Calibration Training Loop}
\label{alg:cpsc}
\begin{algorithmic}[1]
\FOR{epoch = 1 to $E$}
\FOR{batch $\mathcal{B}$ in $\mathcal{D}_{train}$}
    \STATE \texttt{\textcolor{blue}{\% Representation Self-Calibration}}
    \STATE Extract and decompose features $\{h^m\} \rightarrow \{c^m_k\}$
    \STATE Compute reliability scores $\{r^m_k\}$ via CP model
    \STATE Reconstruct $\tilde{h}^m$ from top-$K$ components
    
    \STATE \texttt{\textcolor{blue}{\% Gradient Self-Calibration}} 
    \STATE Compute predictions $\hat{y}^{m} = F_{m}(\{\tilde{h}^m\})$
    \STATE Estimate modality-wise reliability $\rho^{m}$
    \STATE Compute classification loss $\mathcal{L}^{mul}_\text{CE}$ and $\mathcal{L}^{m}_\text{CE}$
    \STATE Apply the gradient modulation $\nabla_{\theta} \mathcal{L}^{m}_{GSC}$
\ENDFOR
    \STATE \texttt{\textcolor{blue}{\% Model Parameter Update}}
    \STATE Update $\theta \leftarrow \theta - \eta (\nabla_{\theta}\mathcal{L} + \nabla_{\theta} \mathcal{L}^{m}_{GSC})$
    
    \STATE \texttt{\textcolor{blue}{\% CP Model Update}}
    \STATE Recompute $\hat{q}$ using $\mathcal{D}_{cal}$ and updated $f_{\theta}$
\ENDFOR
\end{algorithmic}
\end{algorithm}

\subsection{Conformal Predictor Updating}

After updating model parameters to $\theta_{t+1}$, we refresh the CP model to maintain accurate predictive uncertainty quantification. 
Specifically, we recompute nonconformity scores for the calibration set using the updated model:
\begin{equation}
    s_i^{t+1} = 1 - f_{\theta_{t+1}}(x_i)_{y_i}, \quad \forall (x_i, y_i) \in \mathcal{D}_{cal},
\end{equation}
where $s_i^{t+1}$ is the updated nonconformity score. 
The conformal quantile is updated accordingly:
\begin{equation}
    \hat{q}_{t+1} = \text{Quantile}(\{s_i^{t+1}\}_{i=1}^{N}, \lceil (N+1)(1-\alpha) \rceil / N), 
\end{equation}
where $N = |\mathcal{D}_{cal}|$ and $\text{Quantile}(\cdot)$ represent the operation of extraction the quantile at the specified location. 
$\alpha$ is a manual risk factor. 
This updating strategy ensures that the CP model remains synchronized with the current state of $f_{\theta}$, providing relevant guidance for the next training iteration.

After updating the conformal predictor, the overall training loop start the next iteration. 
For clarity, we summarized the self-calibration training loop of our CPSC approach in Alg. \ref{alg:cpsc}. 
During testing, we use the trained model without RSC to obtain robust predictions. 
The features from each unimodal encoders are not required to be decomposed during inference. 
In summary, our CPSC approach only performs during the training stage, thus providing a model-agnostic and unified framework for multimodal learning.

\begin{table*}[thb]
    \centering
    \caption{Comparisons with recent state-of-the-art methods under the Imbalanced Multimodal Learning settings. Note that $Acc_\text{\{m, a, v\}}$ denote multimodal, audio, and visual classification performance. }
        \resizebox{\textwidth}{!}{
        \begin{tabular}{c|cccc|cccc|cccc}
            \hline
            \multirow{2}{*}{Method} & 
      \multicolumn{4}{c|}{Kinetics Sounds} & 
      \multicolumn{4}{c|}{CREMA-D}  & 
      \multicolumn{4}{c}{AVE}\\
      \cline{2-13}
      ~ & $\text{Acc}_\text{m}$ & $\text{Acc}_\text{a}$ & $\text{Acc}_\text{v}$ & Avg & $\text{Acc}_\text{m}$ & $\text{Acc}_\text{a}$ & $\text{Acc}_\text{v}$ & Avg & $\text{Acc}_\text{m}$ & $\text{Acc}_\text{a}$ & $\text{Acc}_\text{v}$ & Avg\\
      \hline
      
      ReconBoost (ICML'24) \cite{reconboost_ICML24} & 70.85 & 56.23 & 50.27 & 59.12 & 79.82 & 60.23 & 73.01 & 71.02 & 71.35 & 61.20 & 39.06 & 57.20 \\
      MMPareto (ICML'24) \cite{Mmpareto_ICML24} & 70.13 & 56.40 & 53.05 & 59.86 & 78.53 & 67.38 & 70.26 & 72.06 & 75.81 & 64.34 & 45.39 & 61.85 \\
      LFM (NeurIPS'24) \cite{LFM_nips2024} & 72.53 & 57.98 & 56.43 & 62.31 & {86.02} & 66.53 & 75.27 & 75.94 & 68.58 & {64.35} & 44.89 & 59.27 \\
      
      InfoReg (CVPR'25) \cite{InfoReg_CVPR2025} & 72.00 & 57.21 & 53.57 & 60.93 & 76.28 & 64.19 & 70.62 & 70.36 & 74.19 & 63.78 & 42.54 & 60.17 \\
    IPRM (IJCAI'25) \cite{IPRM_IJCAI2025} & 74.82 & 59.76 & 58.34 & 64.31 & 85.35 & 65.28 & 76.41 & 75.68 & 74.61 & 65.11 & 43.89 & 61.20 \\
      ARL (ICCV'25) \cite{ARL_ICCV2025} & 74.38 & 58.26 & 59.74 & 64.12 & 83.79 & 65.92 & 72.18 & 73.96 & 70.42 & 63.78 & 39.61 & 57.94 \\
      DGL (ICCV'25) \cite{DGL_ICCV2025} & 74.78 & 52.89 & 60.11 & 62.59 & 82.52 & 65.36 & 74.84 & 74.24 & 73.89 & 64.30 & 42.16 & 60.11 \\
      \hline
      \textbf{CPSC (Ours)} & \textbf{76.08} & \textbf{61.54} & \textbf{61.83} & \textbf{66.48} & \textbf{87.83} & \textbf{67.74} & \textbf{80.38} & \textbf{78.65} & \textbf{77.66} & \textbf{66.93} & \textbf{45.65} & \textbf{63.41} \\
      \hline
    \end{tabular}
    }
    \label{tab:iml}
    \vspace{-4mm}
\end{table*}

\section{Experiments}

\subsection{Experimental Settings}

\noindent
\textbf{Datasets and Metrics.} 
We evaluate our CPSC method on six multimodal datasets, three imbalanced multimodal datasets: CREMA-D \cite{cao2014crema}, AVE \cite{tian2018audio}, Kinetics Sounds \cite{kay2017kinetics}, and three noisy multimodal datasets: SUN RGB-D \cite{song2015sun}, NYU Depth V2 \cite{silberman2012indoor} and MVSA-Single \cite{niu2016sentiment}.
In sum, these datasets cover four modalities including RGB image, depth, audio, and text. 
We evaluate the model under two settings. 
For the imbalanced settings \cite{Mmpareto_ICML24}, we report the multimodal and unimodal accuracy on CREMA-D, AVE, and Kinetics-Sounds to assess the balanced use of modalities. 
For the robustness settings \cite{QMF_ICML23}, we report the multimodal accuracy on SUN RGB-D, NYU Depth V2, and MVSA under synthetic noise corruptions like Gaussian and Salt-Pepper.

\noindent
\textbf{Implementation Details.}
We followed the former works \cite{LFM_nips2024,QMF_ICML23} on imbalanced or robust multimodal learning for fairness. 
Concretely, for audio-visual datasets, the acoustic modality is transformed into 257 $\times$ 1004 spectrograms, and we randomly sample multiple frames from 10-frame video clips. 
The balance factors $\lambda_1$ and $\lambda_2$ are set to 0.8 and 0.2. 
For the RGB-Depth datasets SUN RGB-D and NYU Depth V2, we similarly adopted ResNet18 as the backbone to extract features from both the RGB image and the depth image. 
For the text-image dataset MVSA, ResNet152 was utilized for image feature extraction, while the textual content was encoded using a pre-trained BERT model to obtain sentence-level representations.
\textit{Please refer to our supplementary materials for more details.}

\subsection{Model Comparisons}
We compare CPSC against recent state-of-the-art methods in both imbalanced and robust multimodal learning settings. 
For imbalanced settings, we compare our proposed CPSC method with recent state-of-the-art works including ReconBoost \cite{reconboost_ICML24}, MMPareto \cite{Mmpareto_ICML24}, LFM \cite{LFM_nips2024}, InfoReg \cite{InfoReg_CVPR2025}, DGL \cite{DGL_ICCV2025}, ARL \cite{ARL_ICCV2025}, and IPRM \cite{IPRM_IJCAI2025}. 
For robust settings, we compare our method with baseline methods EAU \cite{EAU_CVPR2024}, ECML \cite{ECML_AAAI24}, and NLC \cite{NLC_AAAI25}. 
Particularly, as a few methods \cite{InfoReg_CVPR2025,DGL_ICCV2025, ARL_ICCV2025, IPRM_IJCAI2025} did not report their complete performance, we reproduce these methods with their official code under the same experimental settings for fair comparisons.

\begin{table}[htbp]
    \centering
    \caption{Comparative results under the Robust Multimodal Learning settings. Note that $\epsilon$ is the noise strength.}
    \label{tab:robustness_comparison}
        \resizebox{\linewidth}{!}{
        \begin{tabular}{c|c|cc|cc}
        \hline
        \multicolumn{6}{c}{MVSA-Single} \\
        \hline
        \multirow{2}{*}{Method} & \multirow{2}{*}{Clean} & \multicolumn{2}{c|}{Gaussian@$\epsilon$} & \multicolumn{2}{c}{Salt-Pepper@$\epsilon$} \\
        \cline{3-4} \cline{5-6}
         ~ & ~ & 5.0 & 10.0 & 5.0 & 10.0 \\
        \hline
        EAU (CVPR'24) \cite{EAU_CVPR2024} & 79.15 & 73.34 & 61.78 & 73.69 & 60.46 \\
        ECML (AAAI'24) \cite{ECML_AAAI24} & 76.83 & 71.28 & 61.03 & 72.13 & 61.04 \\
        NLC (AAAI'25) \cite{NLC_AAAI25} & 73.79 & 65.39 & 58.98 & 66.64 & 57.28 \\
        IPRM (IJCAI'25) \cite{IPRM_IJCAI2025} & 75.84 & 71.25 & 60.69 & 70.13 & 58.26    \\
        ARL (ICCV'25) \cite{ARL_ICCV2025} & 75.76 & 70.89 & 60.74 & 70.49 & 59.82 \\
    
        \hline
        \textbf{CPSC (Ours)} & \textbf{80.07} & \textbf{74.12} & \textbf{63.32} & \textbf{73.95} & \textbf{61.27} \\
        \hline
        \hline
        \multicolumn{6}{c}{NYU Depth V2} \\
        \hline
        \multirow{2}{*}{Method} & \multirow{2}{*}{Clean} & \multicolumn{2}{c|}{Gaussian@$\epsilon$} & \multicolumn{2}{c}{Salt-Pepper@$\epsilon$} \\
        \cline{3-4} \cline{5-6}
         ~ & ~ & 5.0 & 10.0 & 5.0 & 10.0 \\
        \hline
        ECML (AAAI'24) \cite{ECML_AAAI24} & 71.72 & 62.08 & 54.58 & 57.57 & 44.93 \\
        EAU (CVPR'24) \cite{EAU_CVPR2024} & 72.05 & 62.54 & 56.23 & 58.44 & 46.21 \\
        NLC (AAAI'25) \cite{NLC_AAAI25}& 67.33 & 54.90 & 45.04 & 56.02 & 44.66 \\
        IPRM (IJCAI'25) \cite{IPRM_IJCAI2025} & 70.13 & 58.16 & 52.69 & 56.82 & 43.37 \\
        ARL (ICCV'25) \cite{ARL_ICCV2025} & 68.72 & 56.93 & 49.81 & 57.13 & 44.25 \\
        \hline
        \textbf{CPSC (Ours)} & \textbf{73.12} & \textbf{64.15} & \textbf{57.32} & \textbf{61.22} & \textbf{47.40} \\
        \hline
        \hline
        \multicolumn{6}{c}{SUN RGB-D} \\
        \hline
        \multirow{2}{*}{Method} & \multirow{2}{*}{Clean} & \multicolumn{2}{c|}{Gaussian@$\epsilon$} & \multicolumn{2}{c}{Salt-Pepper@$\epsilon$} \\
        \cline{3-4} \cline{5-6}
         ~ & ~ & 5.0 & 10.0 & 5.0 & 10.0 \\
        \hline
        ECML (AAAI'24) \cite{ECML_AAAI24} & 59.82 & 52.46 & 46.71 & 51.92 & 39.29 \\
        EAU (CVPR'24) \cite{EAU_CVPR2024} & 55.68 & 49.39 & 44.23 & 50.38 & 38.38 \\
        NLC (AAAI'25) \cite{NLC_AAAI25} & 52.75 & 43.57 & 38.49 & 45.07 & 37.25 \\
        IPRM (IJCAI'25) \cite{IPRM_IJCAI2025} & 58.67 & 50.82 & 45.16 & 49.98 & 37.29       \\
        ARL (ICCV'25) \cite{ARL_ICCV2025} & 56.29 & 50.78 & 45.92 &  50.34 & 38.46 \\
        \hline
        \textbf{CPSC (Ours)} & \textbf{62.12} & \textbf{54.11} & \textbf{49.10} & \textbf{53.37} & \textbf{41.28} \\
        \hline
        \end{tabular}
        }
        \vspace{-4mm}
    \end{table}

\noindent
\textbf{Evaluation on Imbalanced Multimodal Learning.}
As shown in Table \ref{tab:iml}, our proposed method outperforms all compared approaches on all three audio-visual datasets in terms of both multimodal and unimodal accuracy. 
In particular, compared to the recent state-of-the-art method ARL \cite{ARL_ICCV2025}, our proposed CPSC method shows superior performance across all three benchmarks. 
On the AVE dataset, our CPSC method achieves a 5\% improvement in multimodal accuracy over ARL, while on CREMA-D, the advantage reaches 3\%. 
These results confirm that CPSC can identify and select the most reliable feature components from each modality via the conformal predictor for fusion. 
This is because when dealing with modality imbalance, our CPSC method preserves informative yet scarce patterns from the weaker modalities, preventing useful signals from being overwhelmed by the dominant modality.

\noindent
\textbf{Evaluation on Robust Multimodal Learning.}
According to the results in Table \ref{tab:robustness_comparison}, we list following observations:
(1) our proposed CPSC method demonstrates consistent superiority over existing approaches under both clean and noisy conditions. 
On clean data without corruption, CPSC achieves the best performance on all three datasets. 
This indicates that the self-calibration mechanism not only enhances robustness but also significantly improves fundamental fusion performance on high-quality data by learning more discriminative feature representations.
(2) Under noisy conditions, as the noise intensity increases, all baseline methods exhibit performance degradation, while CPSC maintains stronger stability. 
We speculate the reason is our RSC module selectively suppresses contaminated feature components, while the GSC module automatically reduces the optimization weight of unreliable samples, collectively enhancing the model robustness in noisy environments.

\begin{table}[htbp]
    \centering
    \caption{Ablation study on Imbalanced Multimodal Learning and Robust Multimodal Learning settings. Experiments were conducted on CREMA-D and NYU Depth V2 datasets respectively.}
    \label{tab:ablation}
    \resizebox{0.98\linewidth}{!}{
    \begin{tabular}{cc|cccc}
    \hline
    \multicolumn{2}{c|}{Module} & \multicolumn{4}{c}{Imbalanced Multimodal Learning} \\
    \cline{1-6}
    RSC & GSC & Multiple & Audio & Video & Average \\
    \hline
    $\times$ & $\times$ & 78.23 & 65.41 & 75.66 & 73.10 \\
    $\times$ & $\checkmark$ & 86.29 & 66.68 & 76.28 & 75.32 \\
    $\checkmark$ & $\times$ & 86.35 & 66.55 & 76.72 & 76.20 \\
    $\checkmark$ & $\checkmark$ & 87.83 & 67.74 & 80.38 & 78.65 \\
    \hline
    \end{tabular}
        }
    
    \vspace{1mm}
    
    \resizebox{\linewidth}{!}{
    \begin{tabular}{cc|ccccc}
    \hline
    \multicolumn{2}{c|}{Module} & \multicolumn{5}{c}{Robust Multimodal Learning} \\
    \cline{1-7}
    \multirow{2}{*}{RSC} & \multirow{2}{*}{GSC} & \multirow{2}{*}{Clean} & \multicolumn{2}{c}{Gaussian} & \multicolumn{2}{c}{Salt-Pepper} \\
    \cline{4-5} \cline{6-7}
    & & & 5.0 & 10.0 & 5.0 & 10.0 \\
    \hline
    $\times$ & $\times$ & 69.16 & 59.63 & 51.99 & 56.27 & 41.22 \\
    $\times$ & $\checkmark$ & 70.14 & 59.94 & 50.17 & 55.37 & 41.36 \\
    $\checkmark$ & $\times$ & 71.36 & 63.91 & 53.82 & 61.16 & 45.94 \\
    $\checkmark$ & $\checkmark$ & 73.12 & 64.15 & 57.32 & 61.22 & 47.40 \\
    \hline
    \end{tabular}
    }
    \vspace{-2mm}
    \end{table}

    \begin{figure}[htb]
        \centering
        \centering
        \includegraphics[width=\linewidth]{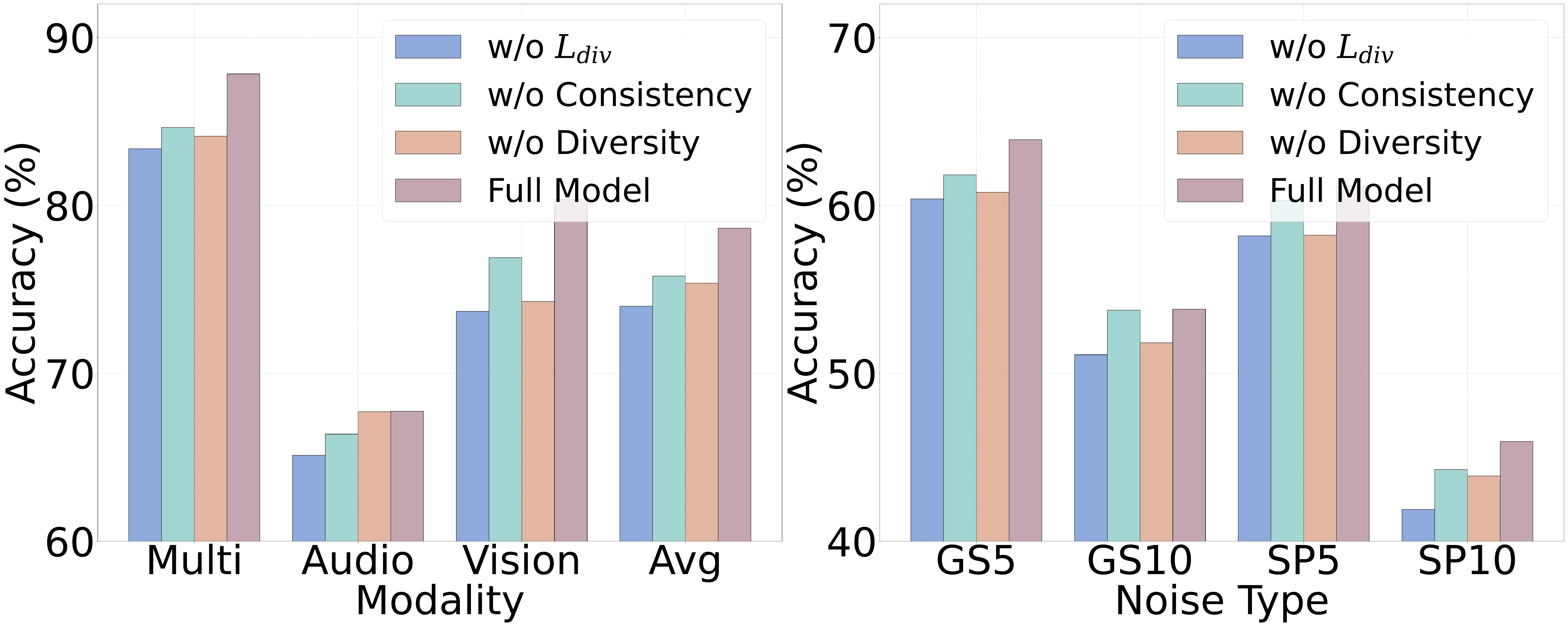}
        \caption{Analysis of the RSC module on the CREMA-D (Left) and NYU Depth V2 (Right) datasets. \textit{GS} and \textit{SP} denoted Gaussian and Salt-Pepper noise respectively. } 
        \label{fig:rsc}
    \vspace{-4mm}
    \end{figure}

\subsection{Further Analysis}

\noindent
\textbf{Ablation Studies.}
To validate the effectiveness of our proposed RSC and GSC modules, we conduct systematic ablation studies. 
From the experimental results in Table \ref{tab:ablation}, we can observe that:  
(1) On the CREMA-D dataset, the full model incorporating both two modules achieves the best performance, significantly outperforming the baseline model and variants with individual modules. 
It demonstrates that the RSC and GSC modules effectively alleviate the modality imbalance problem through distinct yet complementary mechanisms.
(2) Under noisy conditions, the model incorporating RSC achieves a 2-5\% gain over the baseline, whereas the variant with GSC alone shows negligible improvement. 
It indicates that feature-level reliability filtering plays a more crucial role than gradient-level reweighting when explicit noise interference is present.
As the RSC module can directly suppress noise-contaminated feature components, thus preserving model performance.

\noindent
\textbf{Analysis on Representation Self-Calibration.}
To deeply analyze the individual contributions and synergistic effects of the consistency and diversity constraints in our RSC module, we conduct a dedicated ablation study. 
Concretely, we consider four configurations through controlled variable testing: 
the baseline without constraints (w/o $L_\text{div}$), consistency constraint without diversity term (w/o Diversity), diversity constraint without consistency term (w/o Consistency), and full constraints (Full Model). 
According to Fig. \ref{fig:rsc}, we can see that both individual constraints improve model performance, while the full constraints achieve the optimal results. 
This demonstrates that consistency and diversity constraints enhance feature representations through distinct mechanisms: the consistency constraint preserves semantic coherence by preventing feature variants from deviating from the original feature core, while the diversity constraint encourages comprehensive feature space coverage by reducing redundancy among variants.

\begin{table}[htbp]
\centering
\caption{Performances with different optimizers on Kinetics Sounds (KS) and CREMA-D (CD) datasets.}
\label{tab:optimizer_performance}
\resizebox{\linewidth}{!}{
\begin{tabular}{c|ccccc}
\hline
Dataset & Optimizer & $\text{Acc}_m$ & $\text{Acc}_a$ & $\text{Acc}_v$ & $\text{Avg.}$ \\
\hline
\multirow{6}{*}{CD} & SGD & 78.23 & 65.41 & 75.66 & 73.10 \\
~ & SGD+Ours & \textbf{87.83} & \textbf{67.74} & \textbf{80.38} & \textbf{78.65} \\
\cline{2-6}
~ & Adam & 83.78 & 61.86 & 73.78 & 73.14 \\
~ & Adam+Ours & \textbf{85.29} & \textbf{65.38} & \textbf{76.97} & \textbf{75.88} \\
\cline{2-6}
~ & AdaGrad & 68.31 & 56.49 & 34.81 & 53.20 \\
~ & AdaGrad+Ours & \textbf{76.68} & \textbf{59.63} & \textbf{42.18} & \textbf{59.50} \\
\hline
\multirow{6}{*}{KS} & SGD & 67.82 & 51.98 & 42.37 & 54.06 \\
~ & SGD+Ours & \textbf{76.08} & \textbf{61.54} & \textbf{61.83} & \textbf{66.48} \\
\cline{2-6}
~ & Adam & 72.18 & 53.96 & 59.31 & 61.82 \\
~ & Adam+Ours & \textbf{75.13} & \textbf{59.82} & \textbf{62.17} & \textbf{65.71} \\
\cline{2-6}
~ & AdaGrad & 59.48 & 46.58 & 27.96 & 44.67 \\
~ & AdaGrad+Ours & \textbf{65.63} & \textbf{53.27} & \textbf{34.69} & \textbf{51.20} \\
\hline
\end{tabular}
}
\end{table}

\noindent
\textbf{Analysis on Gradient Self-Calibration.}
To verify that our proposed GSC module is not dependent on a specific optimization algorithm, we conduct experiments using multiple adaptive optimizers, including SGD \cite{sgd_2012}, Adam \cite{adam2014method}, and AdaGrad \cite{duchi2011adaptive}. 
According to the comparative results in Table \ref{tab:optimizer_performance}, our method yields significant performance gains across all optimizers. 
On the CREMA-D dataset, SGD integrated with our approach achieves over 10\% improvement in multimodal accuracy, while Adam and the weaker AdaGrad optimizer exhibit gains of approximately 2\% and over 8\%, respectively. 
These results strongly demonstrate that our method effectively mitigates modality imbalance, leading to a more stable optimization process. 
The consistent improvements indicate that our approach successfully guides the optimization trajectory away from local minima, enabling convergence to superior solutions.

\begin{figure}[htb]
    \centering
    \includegraphics[width=\linewidth]{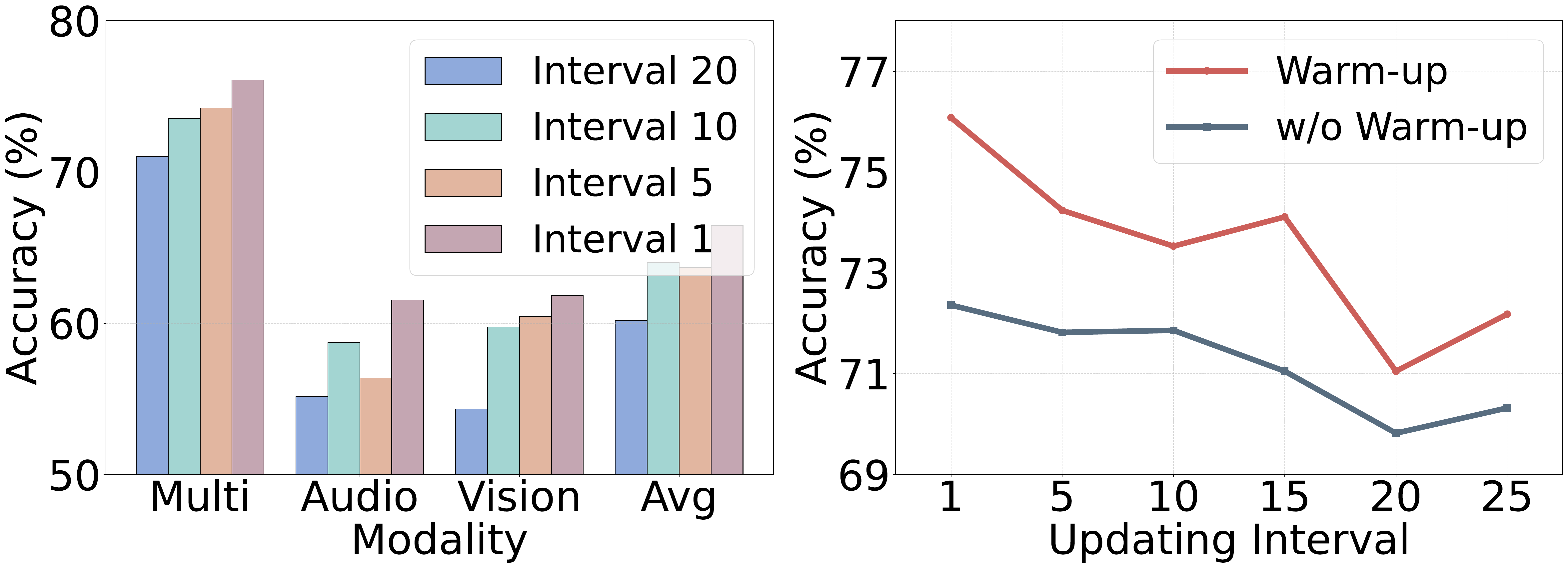}
    \vspace{-6mm}
    \captionof{figure}{Analysis on the conformal predictor updating frequency on Kinetics Sounds datasets. } 
    \label{fig:cp_update}
    \vspace{-2mm}
\end{figure}

\begin{figure}[htb]
    \centering
    \includegraphics[width=0.98\linewidth]{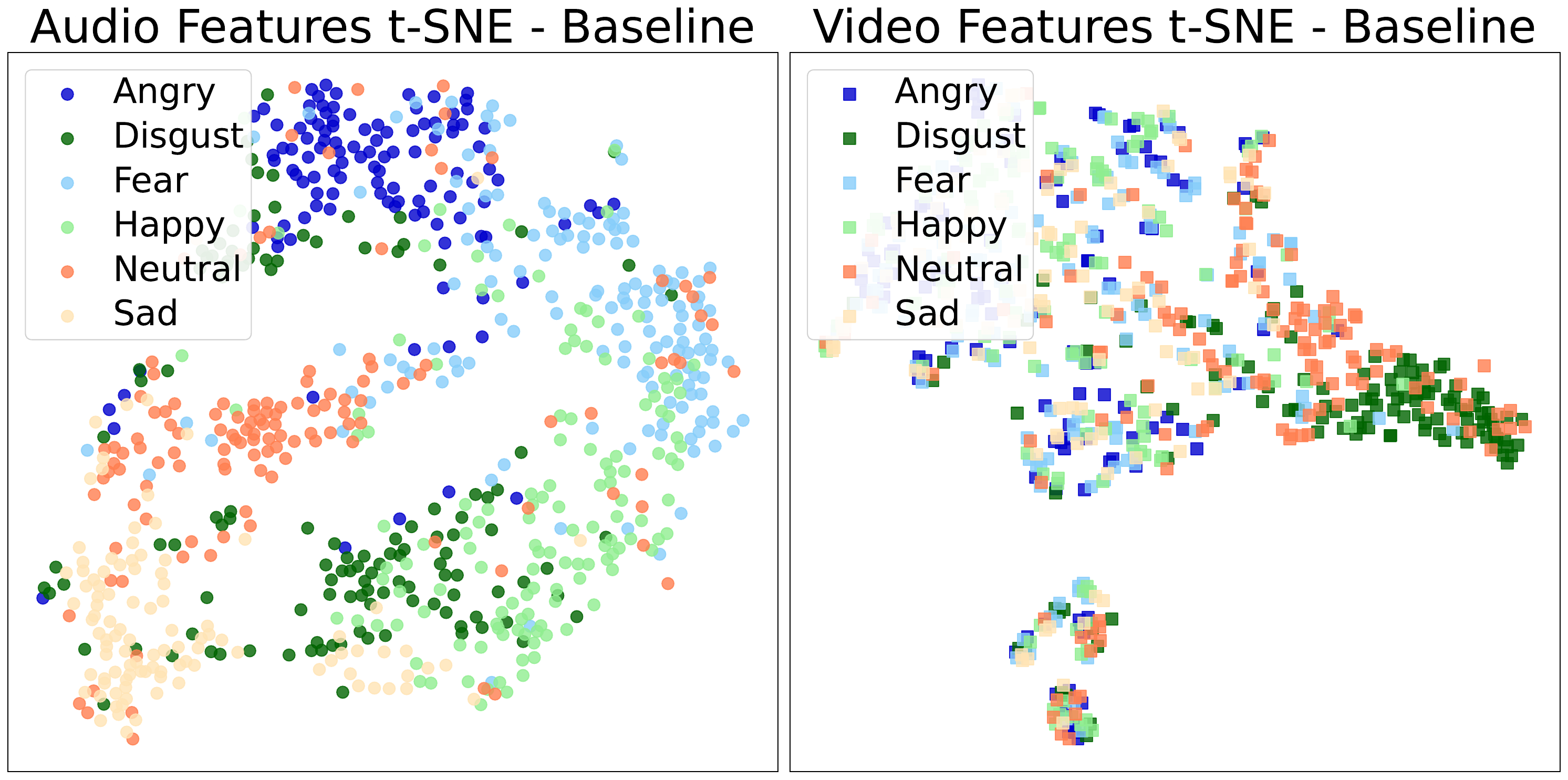}
    \includegraphics[width=0.98\linewidth]{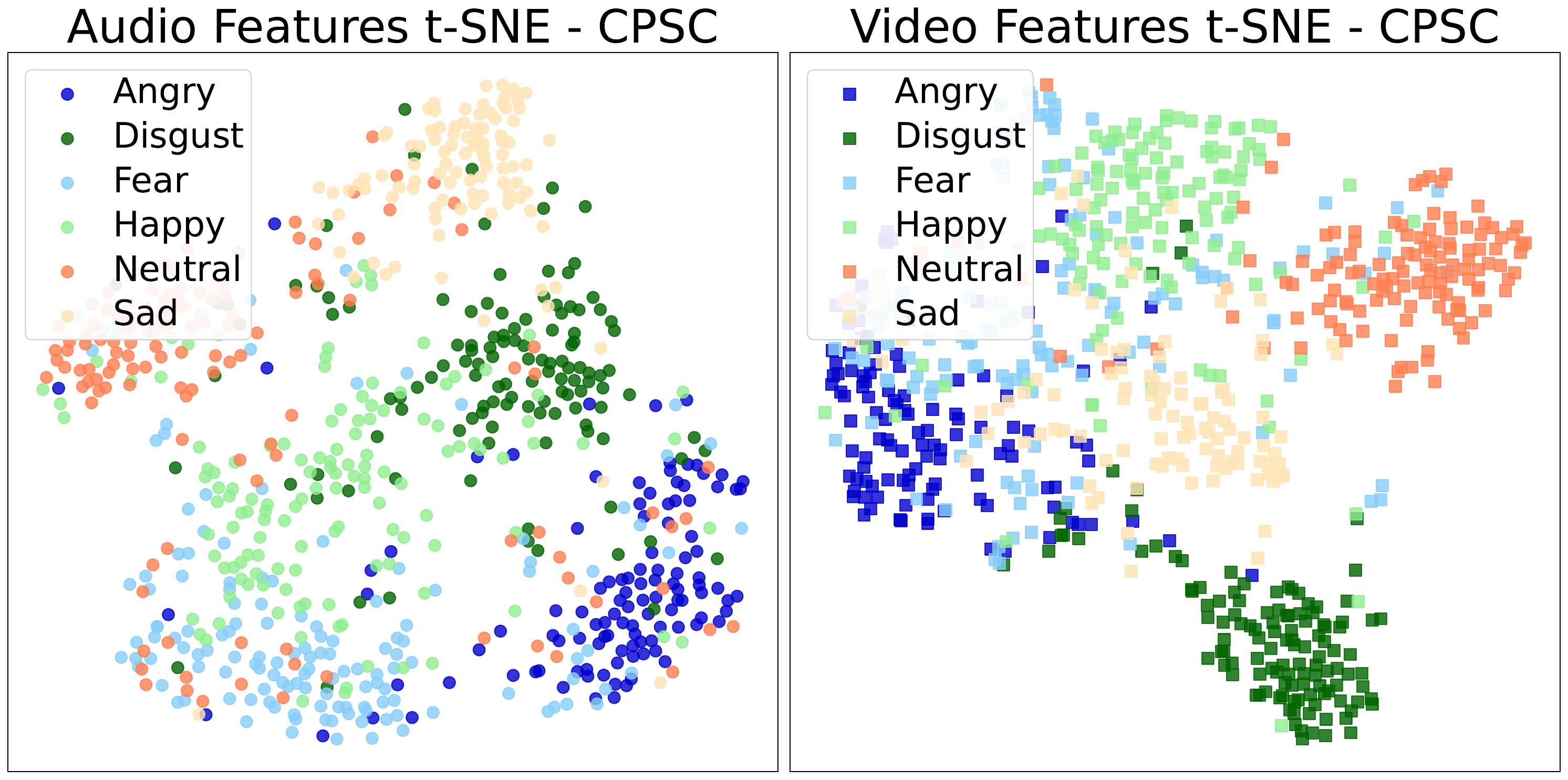}
    \vspace{-2mm}
    \captionof{figure}{Comparative visualizations of audio and visual feature representations on CREMA-D datasets.} 
    \vspace{-4mm}
    \label{fig:qa}
\end{figure}

\noindent
\textbf{Analysis on Conformal Predictor Updating.}
To investigate how the update frequency and initialization strategy of the conformal predictor affect the framework's performance, we conduct ablation studies.
By observing the Fig. \ref{fig:cp_update}, we can see that: 
(1) Regarding update frequency, experimental data clearly indicate that model performance monotonically decreases as the update interval increases. 
One probable reason is that excessively long update intervals cause the conformal predictor to generate calibration signals based on outdated model states. 
(2) The ablated model \textit{w/o Warm-Up} performs significantly worse than our complete approach, revealing the importance of calibration timing. 
The warm-up mechanism allows the model to establish basic representational capacity first, ensuring subsequent calibration is built on a relatively reliable foundation.

\noindent
\textbf{Visualizations of Feature Representation.}
We also conduct t-SNE visualization on the visual and acoustic features learned by the baseline method LFM \cite{LFM_nips2024} and our CPSC approach on the CREMA-D dataset, with results shown in Fig. \ref{fig:qa}. 
The visualization demonstrates a notable advantage of our CPSC method in feature learning compared to the baseline. 
In the acoustic modality, features learned by CPSC exhibit more compact intra-class distribution and clearer inter-class separation, while the baseline features show relatively scattered distributions with noticeable class overlap. 
This advantage becomes more pronounced in the visual modality. 
This visualization result provides feature-level evidence that our method effectively filters reliable feature components through our RSC module.

\begin{figure}[htb]
    \centering
        \centering
        \includegraphics[width=0.95\linewidth]{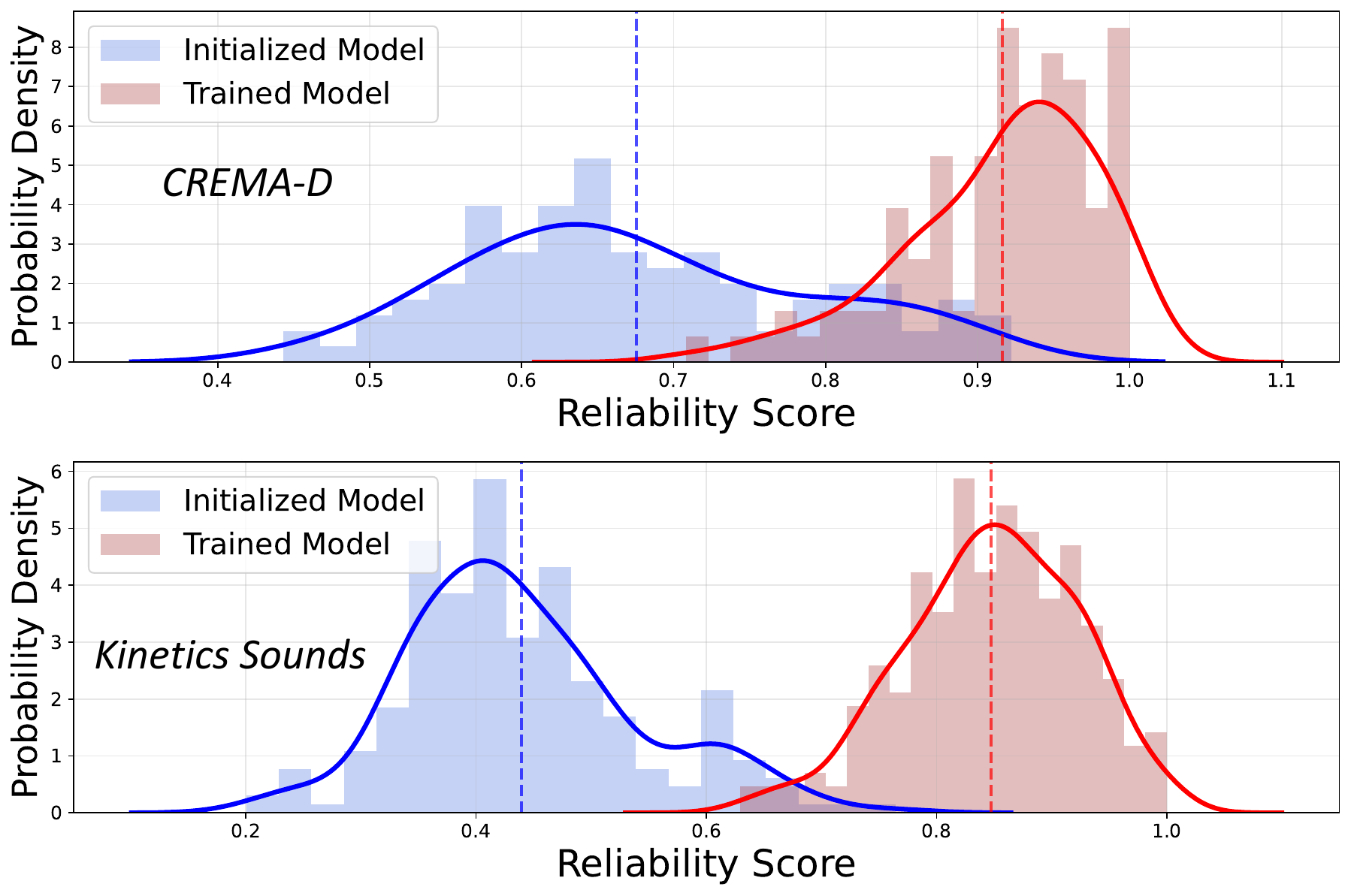}
        \vspace{-2mm}
        \caption{Analysis of reliability of initialized and trained models on the KineticsSounds and CREMA-D datasets.. }
        \vspace{-2mm}
        \label{fig:reliability}
\end{figure}

\noindent
\textbf{Analysis on Reliability Scoring.}
To validate the enhancement of predictive reliability, we also compare the reliability score distributions of initialized and trained models on both the CREMA-D and Kinetics-Sounds datasets.
As illustrated in Fig. \ref{fig:reliability}, experimental results indicate the models trained with our CPSC method exhibit significant improvements in reliability on both datasets.
The post-training reliability distribution shows a pronounced shift toward higher scores with a more compact profile, indicating higher consistency in predictions across individual samples. 
The significant refinement in the reliability distribution confirms that our method effectively addresses the issue of over-confident yet unreliable predictions, enabling the model to produce more accurate and trustworthy outcomes. 
\textit{Additionally, to show the effectiveness of our proposed CP-based strategy, we also provide more analysis regarding the prediction coverage rate and average set size along training epochs in our supplementary materials.}

\section{Conclusion}
In this paper, we presented CPSC, a novel unified framework for multimodal learning on low-quality data that leveraged conformal prediction to enable self-calibration during training. 
It consists of two key components: the Representation Self-Calibration module which enhances model robustness by identifying and preserving reliable features, and the Gradient Self-Calibration module steering optimization toward trustworthy directions. 
Extensive experiments revealed that our proposed CPSC effectively improved the multimodal learning performance on low-quality data. 
For future work, we will explore extending CPSC to other challenging multimodal scenarios.

\section{Acknowledgment}
This work was supported in part by the National Natural Science Foundation of China under Grant No.62476201, the Fundamental Research Funds for the Central Universities, the Central Guidance on Local Science and Technology Development Fund of Shanghai City (No. YDZX20253100002004), and the New Cornerstone Science Foundation through the XPLORER PRIZE.

{
  \small
  \bibliographystyle{unsrt}
  \bibliography{egbib_short}

@inproceedings{ARL_ICCV2025,
  title={Improving Multimodal Learning via Imbalanced Learning},
  author={Wei, Shicai and Luo, Chunbo and Luo, Yang},
  booktitle={ICCV},
  pages={2250--2259},
  year={2025}
}

@InProceedings{DGL_ICCV2025,
    author    = {Wei, Shicai and Luo, Chunbo and Luo, Yang},
    title     = {Boosting Multimodal Learning via Disentangled Gradient Learning},
    booktitle = {ICCV},
    year      = {2025},
    pages     = {22879-22888}
}

@inproceedings{GGDM_MM2025,
  title={Geometric Gradient Divergence Modulation for Imbalanced Multimodal Learning},
  author={Hu, Disen and Jiang, Xun and Sun, Zhe and Yang, Hao and Peng, Chong and Yan, Peng and Shen, Heng Tao and Xu, Xing},
  booktitle={ACM MM},
  pages={1337--1345},
  year={2025}
}

@inproceedings{InfoReg_CVPR2025,
  title={Adaptive unimodal regulation for balanced multimodal information acquisition},
  author={Huang, Chengxiang and Wei, Yake and Yang, Zequn and Hu, Di},
  booktitle={CVPR},
  pages={25854--25863},
  year={2025}
}

@article{LFM_nips2024,
  title={Empowering visible-infrared person re-identification with large foundation models},
  author={Hu, Zhangyi and Yang, Bin and Ye, Mang},
  journal={NeurIPS},
  volume={37},
  pages={117363--117387},
  year={2024}
}

@article{Mmpareto_ICML24,
  title={Mmpareto: Boosting multimodal learning with innocent unimodal assistance},
  author={Wei, Yake and Hu, Di},
  journal={ICML},
  year={2024}
}

@article{reconboost_ICML24,
  title={Reconboost: Boosting can achieve modality reconcilement},
  author={Hua, Cong and Xu, Qianqian and Bao, Shilong and Yang, Zhiyong and Huang, Qingming},
  journal={ICML},
  year={2024}
}

@article{ml_1,
  title={A survey of multimodal learning: Methods, applications, and future},
  author={Yuan, Yuan and Li, Zhaojian and Zhao, Bin},
  journal={ACM Comput. Surv.},
  volume={57},
  number={7},
  pages={1--34},
  year={2025},
}

@article{ml_2,
  title={Foundations \& trends in multimodal machine learning: Principles, challenges, and open questions},
  author={Liang, Paul Pu and Zadeh, Amir and Morency, Louis-Philippe},
  journal={ACM Comput. Surv.},
  volume={56},
  number={10},
  pages={1--42},
  year={2024},
}

@inproceedings{EAU_CVPR2024,
  title={Embracing unimodal aleatoric uncertainty for robust multimodal fusion},
  author={Gao, Zixian and Jiang, Xun and Xu, Xing and Shen, Fumin and Li, Yujie and Shen, Heng Tao},
  booktitle={CVPR},
  pages={26876--26885},
  year={2024}
}

@article{rml_1,
  title={Multimodal fusion on low-quality data: A comprehensive survey},
  author={Zhang, Qingyang and Wei, Yake and Han, Zongbo and Fu, Huazhu and Peng, Xi and Deng, Cheng and Hu, Qinghua and Xu, Cai and Wen, Jie and Hu, Di and others},
  journal={arXiv preprint arXiv:2404.18947},
  year={2024}
}

@inproceedings{QMF_ICML23,
  title={Provable dynamic fusion for low-quality multimodal data},
  author={Zhang, Qingyang and Wu, Haitao and Zhang, Changqing and Hu, Qinghua and Fu, Huazhu and Zhou, Joey Tianyi and Peng, Xi},
  booktitle={ICML},
  pages={41753--41769},
  year={2023},
}

@inproceedings{iml_1,
  title={Balanced multimodal learning via on-the-fly gradient modulation},
  author={Peng, Xiaokang and Wei, Yake and Deng, Andong and Wang, Dong and Hu, Di},
  booktitle={CVPR},
  pages={8238--8247},
  year={2022}
}

@article{iml_2,
  title={Mmpareto: Boosting multimodal learning with innocent unimodal assistance},
  author={Wei, Yake and Hu, Di},
  journal={ICML},
  year={2024}
}

@inproceedings{iml_3,
  title={Intra-and inter-modal curriculum for multimodal learning},
  author={Zhou, Yuwei and Wang, Xin and Chen, Hong and Duan, Xuguang and Zhu, Wenwu},
  booktitle={ACM MM},
  pages={3724--3735},
  year={2023}
}

@inproceedings{iml_4,
  title={Towards balanced active learning for multimodal classification},
  author={Shen, Meng and Huang, Yizheng and Yin, Jianxiong and Zou, Heqing and Rajan, Deepu and See, Simon},
  booktitle={ACM MM},
  pages={3434--3445},
  year={2023}
}

@article{iml_5,
  title={Improving Multimodal Learning Balance and Sufficiency through Data Remixing},
  author={Ma, Xiaoyu and Chen, Hao and Deng, Yongjian},
  journal={ICML},
  year={2025}
}

@inproceedings{rml_4,
  title={Robust multimodal learning via representation decoupling},
  author={Wei, Shicai and Luo, Yang and Wang, Yuji and Luo, Chunbo},
  booktitle={ECCV},
  pages={38--54},
  year={2024},
  organization={Springer}
}

@article{cp_1,
  title={A tutorial on conformal prediction.},
  author={Shafer, Glenn and Vovk, Vladimir},
  journal={JMLR},
  volume={9},
  number={3},
  year={2008}
}

@article{bayesian_1,
  title={A simple baseline for bayesian uncertainty in deep learning},
  author={Maddox, Wesley J and Izmailov, Pavel and Garipov, Timur and Vetrov, Dmitry P and Wilson, Andrew Gordon},
  journal={NeurIPS},
  volume={32},
  year={2019}
}

@inproceedings{bayesian_2,
  title={A Dynamic Bayesian Network Based Framework for Multimodal Context-Aware Interactions},
  author={Han, Violet Yinuo and Wang, Tianyi and Cho, Hyunsung and Todi, Kashyap and Fernandes, Ajoy Savio and Levi, Andre and Zhang, Zheng and Grossman, Tovi and Ion, Alexandra and Jonker, Tanya R},
  booktitle={IUI},
  pages={54--69},
  year={2025}
}

@article{iml_8,
  title={Multiset feature learning for highly imbalanced data classification},
  author={Jing, Xiao-Yuan and Zhang, Xinyu and Zhu, Xiaoke and Wu, Fei and You, Xinge and Gao, Yang and Shan, Shiguang and Yang, Jing-Yu},
  journal={IEEE TPAMI},
  volume={43},
  number={1},
  pages={139--156},
  year={2019},
}

@article{iml_9,
  title={Learning with privileged multimodal knowledge for unimodal segmentation},
  author={Chen, Cheng and Dou, Qi and Jin, Yueming and Liu, Quande and Heng, Pheng Ann},
  journal={IEEE TMI},
  volume={41},
  number={3},
  pages={621--632},
  year={2021},
}

@article{TMC_TPAMI22,
  title={Trusted multi-view classification with dynamic evidential fusion},
  author={Han, Zongbo and Zhang, Changqing and Fu, Huazhu and Zhou, Joey Tianyi},
  journal={IEEE TPAMI},
  volume={45},
  number={2},
  pages={2551--2566},
  year={2022},
}

@inproceedings{rml_6,
  title={Dynamic evidence decoupling for trusted multi-view learning},
  author={Liu, Ying and Liu, Lihong and Xu, Cai and Song, Xiangyu and Guan, Ziyu and Zhao, Wei},
  booktitle={ACM MM},
  pages={7269--7277},
  year={2024}
}

@article{pu_1,
  title={Ensemble quantile networks: Uncertainty-aware reinforcement learning with applications in autonomous driving},
  author={Hoel, Carl-Johan and Wolff, Krister and Laine, Leo},
  journal={IEEE TITS},
  volume={24},
  number={6},
  pages={6030--6041},
  year={2023},
}

@inproceedings{pu_2,
  title={Uncertainty Quantification of Collaborative Detection for Self-Driving},
  author={Su, Sanbao and Li, Yiming and He, Sihong and Han, Songyang and Feng, Chen and Ding, Caiwen and Miao, Fei},
  booktitle={ICRA},
  year={2023}
}

@inproceedings{pu_7,
  title={Uncertainty-debiased multimodal fusion: Learning deterministic joint representation for multimodal sentiment analysis},
  author={Gao, Zixian and Jiang, Xun and Chen, Hua and Li, Yujie and Yang, Yang and Xu, Xing},
  booktitle={ICME},
  pages={1--6},
  year={2024},
}

@inproceedings{pu_8,
  title={Hyperdimensional uncertainty quantification for multimodal uncertainty fusion in autonomous vehicles perception},
  author={Chen, Luke and Wang, Junyao and Mortlock, Trier and Khargonekar, Pramod and Al Faruque, Mohammad Abdullah},
  booktitle={CVPR},
  pages={22306--22316},
  year={2025}
}

@article{cp_risk_1,
  title={Aligning model properties via conformal risk control},
  author={Overman, William and Vallon, Jacqueline and Bayati, Mohsen},
  journal={NeurIPS},
  volume={37},
  pages={110702--110722},
  year={2024}
}

@article{cp_ano_1,
  title={Uncertainty-aware real-time visual anomaly detection with conformal prediction in dynamic indoor environments},
  author={Saboury, Arya and Uyguroglu, Mustafa Kemal},
  journal={IEEE RAL},
  year={2025},
}

@inproceedings{cp_ano_2,
  title={Conformal graph-level out-of-distribution detection with adaptive data augmentation},
  author={Lin, Xixun and Cao, Yanan and Sun, Nan and Zou, Lixin and Zhou, Chuan and Zhang, Peng and Zhang, Shuai and Zhang, Ge and Wu, Jia},
  booktitle={WWW},
  pages={4755--4765},
  year={2025}
}

@article{cp_llm_1,
  title={Analyzing Uncertainty of LLM-as-a-Judge: Interval Evaluations with Conformal Prediction},
  author={Sheng, Huanxin and Liu, Xinyi and He, Hangfeng and Zhao, Jieyu and Kang, Jian},
  journal={EMNLP},
  year={2025}
}

@article{cp_llm_2,
  title={Conformal alignment: Knowing when to trust foundation models with guarantees},
  author={Gui, Yu and Jin, Ying and Ren, Zhimei},
  journal={NeurIPS},
  volume={37},
  pages={73884--73919},
  year={2024}
}

@inproceedings{cp_2,
  title={Conformal Prediction with Learned Features},
  author={Kiyani, Shayan and Pappas, George J and Hassani, Hamed},
  booktitle={ICML},
  pages={24749--24769},
  year={2024},
}

@article{cp_3,
  title={Any2Any: Incomplete Multimodal Retrieval with Conformal Prediction},
  author={Li, Po-han and Yang, Yunhao and Omama, Mohammad and Chinchali, Sandeep and Topcu, Ufuk},
  journal={arXiv preprint arXiv:2411.10513},
  year={2024}
}

@article{cp_4,
  title={Conformal Prediction for Multimodal Regression},
  author={Bose, Alexis and Ethier, Jonathan and Guinand, Paul},
  journal={arXiv preprint arXiv:2410.19653},
  year={2024}
}

@inproceedings{cp_6,
  title={Mutual information-calibrated conformal feature fusion for uncertainty-aware multimodal 3d object detection at the edge},
  author={Stutts, Alex C and Erricolo, Danilo and Ravi, Sathya and Tulabandhula, Theja and Trivedi, Amit Ranjan},
  booktitle={ICRA},
  pages={2029--2035},
  year={2024},
}

@inproceedings{cp_7,
  title={Conformalized multimodal uncertainty regression and reasoning},
  author={Parente, Domenico and Darabi, Nastaran and Stutts, Alex C and Tulabandhula, Theja and Trivedi, Amit Ranjan},
  booktitle={ICASSP},
  pages={6985--6989},
  year={2024},
}

@article{cp_8,
  title={Multi-model Online Conformal Prediction with Graph-Structured Feedback},
  author={Hajihashemi, Erfan and Shen, Yanning},
  journal={TMLR}
}

@inproceedings{NLC_AAAI25,
  title={Noisy label calibration for multi-view classification},
  author={Xu, Shilin and Sun, Yuan and Li, Xingfeng and Duan, Siyuan and Ren, Zhenwen and Liu, Zheng and Peng, Dezhong},
  booktitle={AAAI},
  volume={39},
  number={20},
  pages={21797--21805},
  year={2025}
}

@inproceedings{ECML_AAAI24,
  title={Reliable conflictive multi-view learning},
  author={Xu, Cai and Si, Jiajun and Guan, Ziyu and Zhao, Wei and Wu, Yue and Gao, Xiyue},
  booktitle={AAAI},
  volume={38},
  number={14},
  pages={16129--16137},
  year={2024}
}

@incollection{sgd_2012,
  title={Stochastic gradient descent tricks},
  author={Bottou, L{\'e}on},
  booktitle={Neural networks: tricks of the trade: second edition},
  pages={421--436},
  year={2012},
}

@article{cao2014crema,
  title={Crema-d: Crowd-sourced emotional multimodal actors dataset},
  author={Cao, Houwei and Cooper, David G and Keutmann, Michael K and Gur, Ruben C and Nenkova, Ani and Verma, Ragini},
  journal={IEEE TAFFC},
  volume={5},
  number={4},
  pages={377--390},
  year={2014},
}

@inproceedings{tian2018audio,
  title={Audio-visual event localization in unconstrained videos},
  author={Tian, Yapeng and Shi, Jing and Li, Bochen and Duan, Zhiyao and Xu, Chenliang},
  booktitle={ECCV},
  pages={247--263},
  year={2018}
}

@article{kay2017kinetics,
  title={The kinetics human action video dataset},
  author={Kay, Will and Carreira, Joao and Simonyan, Karen and Zhang, Brian and Hillier, Chloe and Vijayanarasimhan, Sudheendra and Viola, Fabio and Green, Tim and Back, Trevor and Natsev, Paul and others},
  journal={arXiv preprint arXiv:1705.06950},
  year={2017}
}

@inproceedings{song2015sun,
  title={Sun rgb-d: A rgb-d scene understanding benchmark suite},
  author={Song, Shuran and Lichtenberg, Samuel P and Xiao, Jianxiong},
  booktitle={CVPR},
  pages={567--576},
  year={2015}
}

@inproceedings{silberman2012indoor,
  title={Indoor segmentation and support inference from rgbd images},
  author={Silberman, Nathan and Hoiem, Derek and Kohli, Pushmeet and Fergus, Rob},
  booktitle={ECCV},
  pages={746--760},
  year={2012},
  organization={Springer}
}

@inproceedings{niu2016sentiment,
  title={Sentiment analysis on multi-view social data},
  author={Niu, Teng and Zhu, Shiai and Pang, Lei and El Saddik, Abdulmotaleb},
  booktitle={MMM},
  pages={15--27},
  year={2016},
  organization={Springer}
}

@inproceedings{IPRM_IJCAI2025,
  title     = {Towards Equilibrium: An Instantaneous Probe-and-Rebalance Multimodal Learning Approach},
  author    = {Yang, Yang and Wu, Xixian and Jiang, Qing-Yuan},
  booktitle = {IJCAI},
  pages     = {3552--3560},
  year      = {2025},
}

@article{adam2014method,
  title={A method for stochastic optimization},
  author={Adam, Kingma DP Ba J and others},
  journal={ICLR},
  volume={1412},
  number={6},
  year={2014}
}

@article{duchi2011adaptive,
  title={Adaptive subgradient methods for online learning and stochastic optimization.},
  author={Duchi, John and Hazan, Elad and Singer, Yoram},
  journal={JMLR},
  volume={12},
  number={7},
  year={2011}
}

@article{xun_NREP,
  title={Resisting noise in pseudo labels: Audible video event parsing with evidential learning},
  author={Jiang, Xun and Xu, Xing and Zhu, Liqing and Sun, Zhe and Cichocki, Andrzej and Shen, Heng Tao},
  journal={IEEE TNNLS},
  volume={36},
  number={6},
  pages={10874--10888},
  year={2024},
}

@article{xun_GETV,
  title={Generalizable Egocentric Task Verification Via Cross-Modal Hybrid Hypergraph Matching},
  author={Jiang, Xun and Xu, Xing and Wang, Zheng and Song, Jingkuan and Shen, Fumin and Shen, Heng Tao},
  journal={IEEE TPAMI},
  year={2026},
  publisher={IEEE}
}

@inproceedings{shenshen_aaai24,
  title={Adaptive uncertainty-based learning for text-based person retrieval},
  author={Li, Shenshen and He, Chen and Xu, Xing and Shen, Fumin and Yang, Yang and Shen, Heng Tao},
  booktitle={AAAI},
  volume={38},
  number={4},
  pages={3172--3180},
  year={2024}
}

@article{shenshen_tcsvt24,
  title={Cross-modal uncertainty modeling with diffusion-based refinement for text-based person retrieval},
  author={Li, Shenshen and Xu, Xing and He, Chen and Shen, Fumin and Yang, Yang and Shen, Heng Tao},
  journal={IEEE TCSVT},
  volume={35},
  number={3},
  pages={2881--2893},
  year={2024},
}

@article{xun_JOSFD,
  title={Joint objective and subjective fuzziness denoising for multimodal sentiment analysis},
  author={Jiang, Xun and Xu, Xing and Lu, Huimin and He, Lianghua and Shen, Heng Tao},
  journal={IEEE TFS},
  volume={33},
  number={1},
  pages={15--27},
  year={2024},
  publisher={IEEE}
}

@article{wang_TPAMI25,
  title={Evidence-based multi-feature fusion for adversarial robustness},
  author={Wang, Zheng and Xu, Xing and Zhu, Lei and Bin, Yi and Wang, Guoqing and Yang, Yang and Shen, Heng Tao},
  journal={IEEE TPAMI},
  year={2025},
}

@article{xun_ART,
  title={Zero-shot video moment retrieval with angular reconstructive text embeddings},
  author={Jiang, Xun and Xu, Xing and Zhou, Zailei and Yang, Yang and Shen, Fumin and Shen, Heng Tao},
  journal={IEEE TMM},
  volume={26},
  pages={9657--9670},
  year={2024},
}

@inproceedings{hu2026hyper,
  title={Hyper-Opinion Vagueness Quantification for Robust Multimodal Learning},
  author={Hu, Disen and Jiang, Xun and Cao, Xiaofeng and Wang, Zheng and Song, Jingkuan and Shen, Heng Tao and Xu, Xing},
  booktitle={AAAI},
  volume={40},
  number={26},
  pages={21831--21839},
  year={2026}
}
}

\end{document}